\documentclass[11pt]{article}

\usepackage[final]{acl}

\usepackage{times}
\usepackage{latexsym}

\usepackage[T1]{fontenc}

\usepackage[utf8]{inputenc}

\usepackage{microtype}

\usepackage{inconsolata}

\usepackage{graphicx}
\usepackage{amsmath} 
\usepackage{amssymb}
\usepackage{hyperref}
\usepackage{url}
\usepackage{pifont}
\usepackage{xspace}
\usepackage{booktabs}
\usepackage{caption}
\usepackage{multirow}
\usepackage{colortbl}
\usepackage{color}
\usepackage{graphicx}
\usepackage{enumitem}
\usepackage{todonotes}

\newcommand{\method}{\textsc{Forge-RLVR}\xspace}
\newcommand{\data}{\textsc{Forge-Engine}\xspace}

\newcommand{\bench}{\textsc{Forge-Bench}\xspace}

\newcommand{\model}{\textsc{Forge}\xspace}

\newcommand{\heur}{\textsc{Forge-Heuristic}\xspace}

\definecolor{lightgray}{gray}{0.95}
\title{Forge: Quality-Aware Reinforcement Learning for NP-Hard Optimization in LLMs}

\author{ Xiaozhe Li$^{1}$, Xinyu Fang$^{2,}$$^{3}$, Shengyuan Ding$^{2,}$$^{4}$, Yang Li$^{6}$, Linyang Li$^{2}$$^{,5}$, \bf Haodong Duan$^{2}$, \\ \bf Qingwen Liu$^{1}$$^{\dag}$, Kai Chen$^{2}$\\
Tongji University$^{1}$, Shanghai AI Lab$^{2}$, Zhejiang University$^{3}$, Fudan University$^{4}$\\The Chinese University of Hong Kong$^{5}$, Independent $^{6}$ \\
}

\begin{document}
\twocolumn[{%
\renewcommand\twocolumn[1][]{#1}%
\maketitle
\vspace{-20mm}

\begin{center}
    \centering
    \captionsetup{type=figure}
    \vspace{10mm}
    \includegraphics[width=\linewidth]{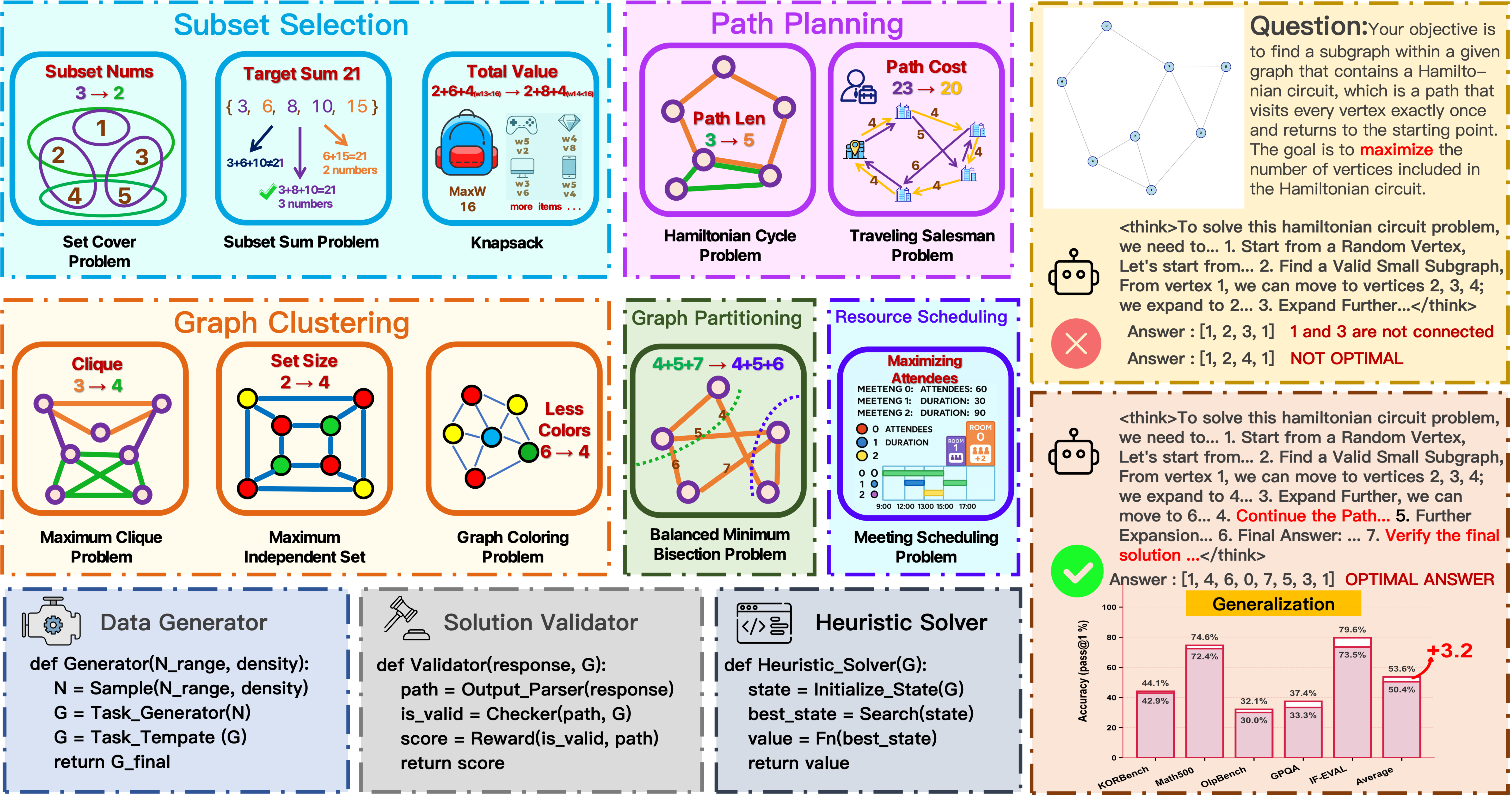}
    \caption{Overview of the \data. The \bench encompasses 10 NP-hard optimization tasks across five categories (e.g., subset selection, path planning), designed to assess reasoning capabilities. An automated pipeline consisting of a Data Generator, Solution Validator, and Heuristic Solver ensures controllable data synthesis, rigorous evaluation, and scalable training. A case study on the Hamiltonian Circuit problem shows the model's capacity to find optimal solutions. Furthermore, evaluations on OOD benchmarks demonstrate that our training enhances general reasoning capabilities.}
    \label{fig:spotlight}
    \vspace{3mm}
\end{center}%
}
]
\footnotetext[1]{$^{\dag}$ represents the corresponding author, email: qliu@tongji.edu.cn.}

\begin{abstract}
Large Language Models (LLMs) have achieved remarkable success on reasoning benchmarks through Reinforcement Learning with Verifiable Rewards (RLVR), excelling at tasks such as math, coding, logic and puzzles. However, existing benchmarks evaluate only correctness, overlooking optimality—the ability to find the best solutions under constraints. We propose \data, the first comprehensive framework for training and evaluating LLMs on NP-hard optimization problems through quality-aware RLVR. \data provides three key components: a scalable training infrastructure with instance generators, quality verifiers, and optimal baselines across 10 tasks; a rigorous benchmark with 1,000 instances evaluating both feasibility (Success Rate) and quality (Quality Ratio); and quality-aware rewards enabling continuous improvement beyond binary correctness. Training on Qwen2.5-7B-Instruct-1M with 15K examples achieves 93.1\% SR and 46.6\% QR, significantly outperforming GPT-4o (29.6\% SR, 14.6\% QR). Beyond optimization, training on \data transfers to diverse tasks: mathematics (+2.2\%), logic (+1.2\%), knowledge (+4.1\%), and instruction-following (+6.1\%). Our analysis reveals quality-aware rewards improve solutions by 28.8\% over binary rewards, and task diversity drives generalization more than data quantity—offering insights into RLVR scaling for complex reasoning.
\end{abstract}

\section{Introduction}

Large Language Models (LLMs) have demonstrated remarkable capabilities on reasoning benchmarks, with recent models achieving near-human performance on mathematics~\cite{he2025deepmath}, coding~\cite{liu2025code}, logic~\cite{xie2025logic}, and puzzle solving~\cite{enigmata, ma2024kor}. These advances have been driven largely by Reinforcement Learning with Verifiable Rewards (RLVR)~\cite{xu2025kodcode, albalak2025big}, which leverages objective, automatically verifiable outcomes to guide model optimization~\cite{o1, guo2025deepseek, gemini, claude}. However, existing reasoning benchmarks focus exclusively on \textit{correctness}—whether answers are right or wrong, true or false, pass or fail. This binary evaluation overlooks a critical dimension: \textbf{optimality}—the ability to find not just valid solutions, but solutions that maximize or minimize an objective function.

In many practical scenarios, finding a \textit{feasible} solution is merely the baseline; the true intelligence lies in finding the \textit{optimal} one. Consider a logistics agent planning a route for 20 cities: while there are $2.4 \times 10^{18}$ valid paths, a route that is merely "valid" but 3$\times$ longer than necessary is functionally useless. This distinction between \textit{correctness} (feasibility) and \textit{optimality} (efficiency) represents a significant gap in current LLM capabilities. Even state-of-the-art models like GPT-4o often settle for the first valid solution they find, struggling to engage in the iterative optimization search required to refine a solution towards optimality.

To bridge this gap, we identify \textbf{NP-hard combinatorial optimization problems} as the ideal testbed for advancing LLM reasoning. These problems offer a unique trifecta of properties perfectly suited for RLVR training. \textbf{First}, they are computationally intractable, forcing the model to develop strategic heuristics rather than relying on rote memorization or brute force. \textbf{Second}, unlike standard logic tasks with sparse binary rewards (0 or 1), NP-hard problems possess an \textit{objective quality score} (e.g., total tour length, subset value). This allows us to construct \textbf{dense, continuous reward signals}—rewarding a solution that is 90\% optimal more than one that is 50\% optimal—thereby guiding the model through a smoother optimization landscape. \textbf{Third}, verification is efficient (polynomial time), enabling scalable, automated supervision without human labeling.

Despite this potential, utilizing NP-hard problems for RLVR remains unexplored. Existing benchmarks like NPHardEval~\cite{nphardeval} and NPPC~\cite{yang2025nppc} focus primarily on feasibility (e.g., "Is there a solution?") or rely on coarse evaluation metrics. Crucially, they lack a mechanism to generate the \textit{ground-truth optimal values} needed to compute fine-grained rewards. Without the optimal solution, an RL system cannot judge the quality of the model's output, reverting the training process back to inefficient binary feedback.

To address this, we propose \method, the first comprehensive framework designed to empower optimization reasoning in LLMs via RLVR. \method encompasses 10 distinct tasks across five domains (e.g., Scheduling, Routing, Clustering). The core innovation lies in our \textbf{Generator-Verifier-Solver pipeline}:
(i) \textbf{Controllable Generators} produce infinite training instances with graded difficulty (Easy/Medium/Hard) to facilitate curriculum learning;
(ii) \textbf{Rule-based Verifiers} ensure strict constraint satisfaction; and most importantly,
(iii) \textbf{Heuristic Solvers} rapidly compute near-optimal baselines for every generated instance.
This pipeline transforms NP-hard problems from simple evaluation tasks into a rich training environment where models receive precise feedback on their \textit{optimality gap}.

We further introduce \bench, a rigorous evaluation suite of 1,000 high-complexity instances. Using this framework, we train \model (based on Qwen2.5-7B) using multi-stage RLVR. The results are striking: our 7B model achieves a 93.1\% Success Rate and 46.6\% Quality Ratio, significantly outperforming GPT-4o (62.1\%SR, 36.2\% QR) on in-domain combinatorial optimization tasks.

Beyond optimization tasks, we investigate whether optimization training transfers to general reasoning. Specifically, we ask: \textit{Does learning to solve the Traveling Salesman Problem improve a model's ability on general math or logic tasks?} The answer is affirmative. \model shows consistent gains across diverse reasoning benchmarks, including Math (+2.2\%), Logic (+1.2\%), Knowledge (+4.1\%), and Instruction Following (+6.1\%). These results suggest that optimization training fosters a transferable \textbf{``optimization mindset''}—the ability to generate candidates, and self-refine—benefiting broader reasoning capabilities.

Our key contributions are:
\begin{itemize}[leftmargin=*,topsep=2pt,itemsep=1pt]
    \item \textbf{\data Framework:} We establish an end-to-end infrastructure for optimization reasoning, featuring the novel integration of heuristic solvers to provide fine-grained optimality rewards. This enables effective RLVR training that is superior to standard SFT.
    
    \item \textbf{\bench Benchmark:} We propose a dual-metric benchmark (Success Rate \& Quality Ratio) that rigorously evaluates the depth of optimization reasoning, moving beyond the binary correctness paradigm of prior works.
    
    \item \textbf{OOD Generalization:} We provide empirical evidence that optimization training transfers to general reasoning capabilities. Our analysis reveals that \textit{task diversity} and \textit{quality-aware rewards} are the critical factors driving this generalization, offering new insights into the scaling behavior of RLVR-based training.
\end{itemize}

\section{\data: Data Construction}
\begin{figure*}[h]
    \centering
    \includegraphics[width=0.95\linewidth]
    {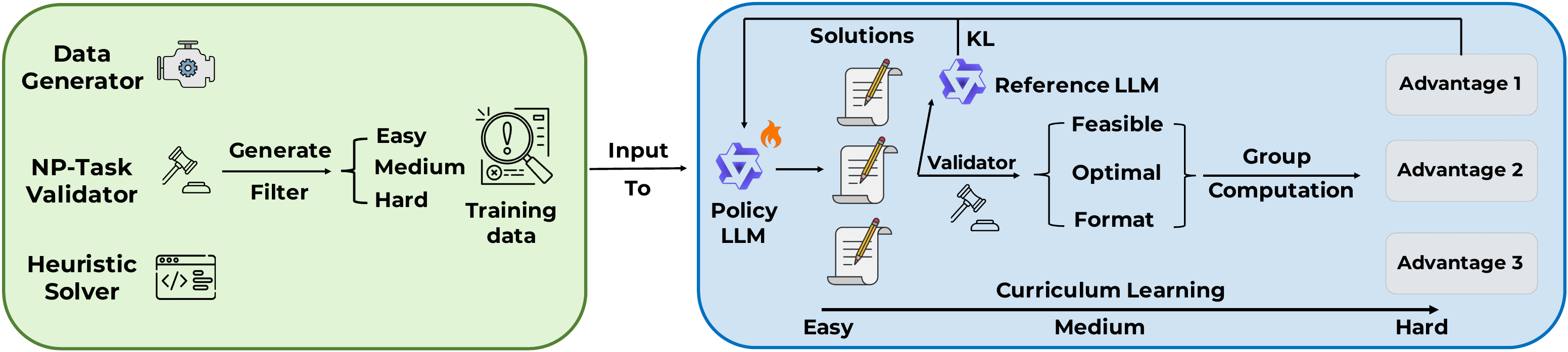}
    \caption{\method training pipeline with quality-aware RLVR. The model generates solutions with step-by-step reasoning, which are evaluated through three components: (i) format verification checking output structure, (ii) feasibility verification ensuring constraint satisfaction, and (iii) quality assessment measuring optimality relative to heuristic baselines. The combined reward signal guides model optimization. Training progresses through curriculum learning across three stages (Easy $\rightarrow$ Medium $\rightarrow$ Hard), enabling progressive skill development from constraint satisfaction to sophisticated optimization.}
    \label{fig:pipeline}
    \vspace{-1em}
\end{figure*}

\data provides comprehensive infrastructure for training and evaluating LLMs on NP-hard optimization. The framework consists of: (i) scalable training infrastructure with generators, verifiers, and optimal baselines across 10 tasks, and (ii) a rigorous benchmark with 1,000 high-complexity instances. We describe the task categories, data construction pipeline, and evaluation protocol below.\looseness-1

\subsection{Task Categories and Problem Domains}

\data covers 10 NP-hard optimization tasks spanning five fundamental categories (Figure~\ref{fig:spotlight}). Each category represents a distinct class of combinatorial optimization problems with different constraint structures and reasoning requirements.

\noindent \textbf{Graph Clustering.}  
These tasks require identifying vertex subsets that satisfy strict adjacency constraints while optimizing structural objectives. We include three canonical problems: \texttt{Maximum Clique} (finding the largest fully-connected subgraph), \texttt{Maximum Independent Set} (finding the largest set of non-adjacent vertices), and \texttt{Graph Coloring} (minimizing the number of colors needed such that no adjacent vertices share a color). These tasks test graph structural understanding, conflict resolution, and chromatic reasoning.

\noindent \textbf{Resource Scheduling.}  
These tasks involve assigning activities to limited resources while avoiding conflicts and maximizing a global objective. The \texttt{Meeting Scheduling} problem requires allocating meetings to rooms and time slots while respecting attendee availability, room capacity, and temporal constraints. This task tests multi-constraint optimization and resource allocation strategies.

\noindent \textbf{Graph Partitioning.}  
These tasks require dividing graphs into balanced subsets while minimizing edge cuts between partitions. \texttt{Balanced Minimum Bisection} partitions a graph into two nearly equal-sized sets while minimizing the total weight of edges crossing the partition. This task tests the ability to balance competing objectives: partition balance versus cut minimization.

\noindent \textbf{Subset Selection.}  
These tasks involve choosing subsets of items under combinatorial constraints to optimize coverage or value-weight trade-offs. We include three representative problems: \texttt{Subset Sum} (selecting elements that sum to a target value while maximizing subset size), \texttt{Set Cover} (covering all elements with minimum number of sets), and \texttt{Knapsack} (maximizing total value subject to weight capacity). These tasks test discrete optimization and constraint satisfaction.

\noindent \textbf{Path Planning.}  
These tasks require finding optimal tours or cycles through all nodes in a graph. \texttt{Traveling Salesman Problem} (TSP) seeks the shortest tour visiting each city exactly once, while \texttt{Hamiltonian Cycle} finds the longest cycle visiting all vertices. These tasks test global optimization, permutation reasoning, and spatial planning.

\subsection{Data Construction Pipeline}

We construct \data through a systematic four-stage pipeline ensuring scalability, verifiability, and controllable difficulty.

\noindent\textbf{Stage I: Task Selection.}  
We select 10 NP-hard tasks spanning diverse constraint types with well-defined objectives and real-world relevance. Each task is formalized with clear input/output specifications and optimization objectives (e.g., TSP: given $n$ cities with pairwise distances, find the shortest tour visiting each city exactly once).

\noindent\textbf{Stage II: Generators and Verifiers.}  
For each task, we develop: 
(i) \textit{Programmable generators} that produce diverse instances with controllable parameters (e.g., complexity levels, graph density). To ensure \textbf{solvability}, we employ a \textbf{construction-by-design} approach, where each instance is synthesized from a known valid solution state to guarantee the existence of at least one feasible solution. 
(ii) \textit{Rule-based verifiers} that automatically validate constraint satisfaction and compute objective values. These verifiers are cross-referenced with ground-truth solutions and have been manually audited to ensure absolute correctness.

\noindent\textbf{Stage III: Heuristic Solvers.}  
We implement task-specific heuristic algorithms that generate high-quality approximate solutions, providing baselines for reward computation during training. Heuristics are designed to be efficient (polynomial time) and effective (near-optimal quality). For example, our TSP heuristic uses multi-start nearest-neighbor initialization followed by 2-opt local search until convergence or timeout.

\noindent\textbf{Stage IV: Difficulty Calibration.}  
We define three difficulty levels—\textit{Easy}, \textit{Medium}, and \textit{Hard}—by systematically varying problem size and complexity parameters. Difficulty levels are calibrated empirically: we generate candidate instances across parameter ranges, evaluate baseline model performance, and select parameter settings that produce target success rates (Easy: 70-90\%, Medium: 40-70\%, Hard: 10-40\%). For example, TSP difficulty is controlled by the number of cities: Easy (10-20 cities), Medium (20-30 cities), Hard (35-45 cities).

\subsection{\bench: Evaluation Benchmark}
\begin{table*}[t]
    \centering
    \small
    \caption{Comparison of \bench with prior reasoning and optimization benchmarks. \bench is the first benchmark that combines scalability (unlimited instance generation), quality-aware evaluation (measuring solution optimality, not just correctness), and RLVR-readiness (fine-grained rewards for training). Existing benchmarks either lack scalable generation, measure only correctness, or are not suitable for RLVR training.}
    \begin{tabular}{lcccccc}
    \toprule
        \textbf{Benchmark} & \textbf{Task Type} & \textbf{Tasks} & \textbf{Scalable} & \textbf{Verifier} & \textbf{Trainable} & \textbf{Quality} \\ 
        \midrule
        KOR-Bench~\cite{ma2024kor} & Knowledge & 125 & \ding{55}  & \ding{51} & \ding{55} & \ding{55}\\ 
        NPR~\cite{wu2025phd} & Knowledge & 1 & \ding{55} & \ding{55} & \ding{55} & \ding{55}\\ 
        Logic-RL~\cite{xie2025logic} & Logic & 1 & \ding{51} & \ding{51} & \ding{51} & \ding{55}\\
        ZebraLogic~\cite{lin2025zebralogic} & Logic & 1 & \ding{51} & \ding{51} & \ding{51} & \ding{55}\\ 
        SearchBench~\cite{borazjanizadeh2024navigating} & Puzzle & 11 & \ding{51} & \ding{51} & \ding{55} & \ding{55}\\ 
        Enigmata~\cite{enigmata} & Puzzle & 36 & \ding{51} & \ding{51} & \ding{51} & \ding{55}\\ 
        NPHardEval~\cite{nphardeval} & NP & 9 & \ding{55}  & \ding{55} & \ding{55} & \ding{55}\\ 
        \midrule
        \bench & NP & 10 & \ding{51} & \ding{51} & \ding{51} & \ding{51}\\ 
        \bottomrule
    \end{tabular}
    \label{tab:comparison}
\end{table*}

Building on \data, we introduce \textbf{\bench}, a rigorous benchmark for evaluating LLM optimization capabilities.

\noindent\textbf{Benchmark Design.}  
\bench contains 1,000 test instances: 100 instances per task, with instances drawn from the high-complexity range to challenge model capabilities. For example, TSP instances contain 45-55 cities, significantly larger than training instances. All test instances are generated using the same generators as training data but with different complexity control variables, ensuring they are unseen during training while maintaining consistent problem structure.

\noindent\textbf{Evaluation Metrics.}  
Unlike existing benchmarks that measure only correctness, \bench assesses \textit{optimization capability} through two complementary metrics:

\textbf{Success Rate (SR):} The percentage of instances for which the model generates a valid solution satisfying all constraints. This measures the model's ability to understand problem constraints and produce feasible solutions.

\textbf{Quality Ratio (QR):} The quality of model solutions compared to heuristic baselines. For minimization problems (e.g., TSP), $\text{QR} = \frac{\text{Heuristic Solution}}{\text{Model Solution}}$; for maximization problems (e.g., Knapsack), $\text{QR} = \frac{\text{Model Solution}}{\text{Heuristic Solution}}$. Invalid solutions receive QR = 0. This metric measures solution quality: QR = 1.0 indicates matching the heuristic baseline, while QR < 1.0 indicates suboptimal solutions.

Together, SR and QR provide a comprehensive assessment: SR measures whether models can generate valid solutions, while QR measures how close these solutions are to optimal. This dual-metric approach distinguishes \bench from prior benchmarks that evaluate only feasibility.

\noindent\textbf{Key Advantages.}  
Table~\ref{tab:comparison} compares \bench with prior benchmarks in the NP-hard and reasoning domains. \bench offers several advantages over existing benchmarks: (i) \textit{Scalability}—generators produce unlimited test instances at any difficulty level; (ii) \textit{Quality-awareness}—QR metric enables fine-grained assessment of solution optimality; (iii) \textit{Automatic evaluation}—verifiers provide instant feedback without human annotation; (iv) \textit{Contamination-free}—all instances are generated procedurally, guaranteeing they do not appear in any training corpus; and (v) \textit{Comprehensive coverage}—10 tasks across 5 categories test diverse optimization capabilities.

\section{\method: Quality-Aware RL}

Standard reasoning RLVR relies on binary feedback (correct/incorrect), which is insufficient for combinatorial optimization where the goal is to maximize solution quality, not just validity. To address this, we propose a \textbf{Quality-Aware RLVR} framework. This approach replaces sparse binary signals with fine-grained, continuous feedback derived from heuristic baselines and employs a difficulty-based curriculum to guide the model from basic constraint satisfaction to high-level optimization strategies.

\subsection{Quality-Aware Reward Function}

As illustrated in Figure~\ref{fig:pipeline}, our reward function evaluates outputs sequentially across three dimensions: format, feasibility, and optimality. This hierarchical design ensures the model first learns to structure its reasoning, then to satisfy constraints, and finally to refine its solutions toward optimality.

\noindent \textbf{Format Reward ($R_{\text{format}}$).} 
To ensure verifiable and interpretable reasoning, we require outputs to follow a strict structure: step-by-step reasoning within \texttt{<think>} tags followed by a clearly delimited answer.
\[
R_{\text{format}} = 
\begin{cases} 
+1, & \text{if format is valid} \\
-1, & \text{if format is invalid}
\end{cases}
\]
\noindent \textbf{Feasibility Reward ($R_{\text{feasibility}}$).}
This component enforces problem-specific hard constraints (e.g., unique city visits in TSP, capacity limits in Knapsack). Invalid solutions receive a penalty, while valid solutions inherit the optimality score:
\[
R_{\text{feasibility}} = 
\begin{cases} 
R_{\text{optimal}}, & \text{if solution is feasible} \\
-1.5, & \text{if solution is infeasible}
\end{cases}
\]
\noindent \textbf{Optimality Reward ($R_{\text{optimal}}$).} 
For feasible solutions, we compute a continuous quality score by comparing the model's objective value ($M_s$) against a heuristic baseline ($M_h$). This provides the dense signal necessary for optimization:
\[
R_{\text{optimal}} = 
\begin{cases} 
M_s / M_h, & \text{maximization tasks} \\
M_h / M_s, & \text{minimization tasks}
\end{cases} 
\]
$R_{\text{optimal}} \in (0, 1]$, where a value closer to 1.0 indicates near-optimal performance. The final total reward is computed as the sum of the structural and feasibility components: $R = R_{\text{format}} + R_{\text{feasibility}}$.

\subsection{Difficulty-Based Curriculum}

Training directly on complex NP-hard instances leads to the "cold start" problem, where rewards are too sparse for effective learning. We mitigate this via a progressive curriculum strategy (Easy $\rightarrow$ Medium $\rightarrow$ Hard). 
In the early phase (Easy), the model focuses on mastering \textit{feasibility}—learning to generate valid structures and satisfy basic constraints. As difficulty increases (Medium/Hard), the focus shifts to \textit{optimality}, challenging the model to refine its strategies within larger search spaces. This structured progression ensures a stable learning trajectory, preventing the model from collapsing under the complexity of hard instances.

\subsection{Multi-Stage Curriculum Replay} 
Standard linear curriculum learning often suffers from catastrophic forgetting, where the model's proficiency on foundational tasks degrades as it adapts to higher complexities. To mitigate this, we propose \textbf{Multi-Stage Curriculum Replay}, a cyclic training strategy that iterates through the difficulty hierarchy (Easy $\rightarrow$ Medium $\rightarrow$ Hard) multiple times. By dividing the training process into distinct stages that each revisit the full difficulty spectrum using fresh instances, this approach continuously reinforces basic optimization reasoning while progressively stabilizing performance on harder problems, ensuring robust generalization across all difficulty levels.

\section{Experiments}


\subsection{Experimental Setup}

\paragraph{Evaluation Benchmarks.} We evaluate on \bench (1,000 instances across 10 NP-hard tasks) and five OOD benchmarks: (1) \textbf{Logic}: KORBench; (2) \textbf{Mathematics}: Math500 and OlympiadBench.; (3) \textbf{Knowledge}: GPQA\_Diamond; (4) \textbf{Instruction-Following}: IFEval. All experiments use OpenCompass~\cite{2023opencompass}.

\paragraph{Training Configuration.} We train \model on Qwen2.5-7B-Instruct-1M~\cite{yang2025qwen251mtechnicalreport} using curriculum replay with three stages (5K instances per stage, 15K total). We use Qwen2.5-7B-Instruct-1M rather than standard Qwen2.5-7B-Instruct as it provides better instruction adherence during RLVR training. We constructed a 10K SFT dataset via distillation from Qwen3-235B-Thinking, applying RFT to select solutions with a Quality Ratio $>0.5$. Additionally, we utilized DAPO-17K~\cite{yu2025dapo17k} for math tasks. Full training details are listed in the Appendix~\ref{apd:setting}.

\subsection{Main Results}

As shown in Tables~\ref{tab:indomain} and~\ref{tab:ood}, \model achieves 93.1\% SR and 46.6\% QR on \bench, tripling the base model's performance and significantly outperforming GPT-4o (62.1\% SR). Crucially, this optimization capability transfers to general reasoning: \model demonstrates consistent gains across logic, math, knowledge, and instruction-following, raising the overall OOD score from 50.4 to 53.6 (+3.2 points).

\paragraph{Performance by Task Category.} 
Performance gains correlate strongly with task complexity. The largest improvements occur in {Graph} and {Planning} domains, where rigid structural constraints (e.g., Hamiltonian cycles) cause base models to fail almost entirely. RLVR closes this gap, achieving near-perfect performance (e.g., 98.2\% SR in Planning). {Selection} tasks also show notable gains. These results highlight RLVR’s effectiveness in strengthening LLM reasoning under high-dimensional, tightly constrained search spaces where base models often fail. For OOD generalization, we observe strong transfer to {Knowledge Reasoning} (+4.1\%) and {Instruction Following} (+6.1\%), suggesting that optimization training fosters broadly applicable reasoning skills.

\paragraph{Joint Training Synergy.} Combining \data with mathematical reasoning data (DAPO-17K) yields further benefits. The joint model (\textbf{FORGE-7B-NP-MATH}) maintains high in-domain success (91.3\% SR) while enhancing solution quality (51.3\% QR). Moreover, it achieves the highest OOD average (54.2), surpassing both NP-only (53.6) and math-only (53.9) baselines. This indicates that optimization and mathematical reasoning are mutually reinforcing: optimization training likely sharpens constraint satisfaction useful for math, while mathematical training strengthens the logical rigor required for high-quality optimization.

\paragraph{Transfer Mechanisms.} We attribute this cross-domain generalization to three cognitive capabilities fostered by optimization training: (1) \textit{Constraint Reasoning}, which aids instruction following; (2) \textit{Quality Evaluation}, which improves error detection; and (3) \textit{Iterative Refinement}, which enhances self-correction. The pronounced gains in instruction-following (+6.1\%) and knowledge reasoning (+4.1\%) provide strong empirical support for this hypothesis.

\begin{table*}[t]
    \caption{Performance of reasoning LLMs, general LLMs, and our trained LLMs on \bench.}
    \centering
    \small
    \resizebox{\textwidth}{!}{
    \begin{tabular}{lcccccccccccc}
    \toprule
    \multirow{2}{*}{\textbf{Model}} &
      \multicolumn{2}{c}{\textbf{Graph}} & \multicolumn{2}{c}{\textbf{Schedule}} & \multicolumn{2}{c}{\textbf{Partition}} & \multicolumn{2}{c}{\textbf{Selection}} & \multicolumn{2}{c}{\textbf{Planning}} & \multicolumn{2}{c}{\textbf{Overall}} \\
    \cmidrule(lr){2-3} \cmidrule(lr){4-5} \cmidrule(lr){6-7} \cmidrule(lr){8-9} \cmidrule(lr){10-11} \cmidrule(lr){12-13}
     & 
      \textbf{SR} & \textbf{QR} & \textbf{SR} & \textbf{QR} & \textbf{SR} & \textbf{QR} & \textbf{SR} & \textbf{QR} & \textbf{SR} & \textbf{QR} & \textbf{SR} & \textbf{QR} \\
    \midrule
    \rowcolor{lightgray}
    \multicolumn{13}{c}{\emph{Proprietary LLMs}} \\ 
    \midrule
    DS-V3.1-Thinking & 86.0 & 78.2 & \textbf{99.0} & 91.4 & 98.0 & \textbf{77.1} & \textbf{99.3} & \textbf{98.5} & 61.9 & 54.3 & 88.8 & \textbf{79.9} \\
    gpt-o3 & \textbf{97.0} & \textbf{86.4} & \textbf{99.0} & 94.5 & \textbf{100.0} & 51.4 & 87.4 & 87.3 & 74.0 & \textbf{65.1} & 91.5 & 76.9 \\
    Qwen3-235B-Thinking & 66.7 & 62.9 & 95.0 & 93.0 & \textbf{100.0} & 55.8 & 98.0 & 97.1 & 52.0 & 44.8 & 82.3 & 70.7 \\
    {Qwen3-235B-Instruct} & 47.0 & 37.9 & 86.0 & 80.0 & 100.0 & 50.4 & 78.7 & 74.4 & 15.5 & 8.3 & 65.4 & 50.2 \\
    gpt-4o-2024-08-06 & 64.7 & 29.3 & 79.0 & 59.8 & \textbf{100.0} & 53.0 & 14.7 & 9.3 & 52.0 & 29.6 & 62.1 & 36.2 \\
    \midrule
    \rowcolor{lightgray}
    \multicolumn{13}{c}{\emph{Open-Source LLMs}} \\ 
    \midrule
    {Qwen3-30B-A3B} & 57.3 & 50.6 & 97.0 & \textbf{97.5} & 99.0 & 51.2 & 86.0 & 84.0 & 44.5 & 36.9 & 76.8 & 64.1 \\
    Qwen3-32B & 44.7 & 39.3 & 94.0 & 93.9 & 99.0 & 52.6 & 94.1 & 91.4 & 21.6 & 11.2 & 70.7 & 57.6 \\
    Qwen3-8B & 22.7 & 16.8 & 78.0 & 75.3 & 98.0 & 51.0 & 86.0 & 82.6 & 3.0 & 1.2 & 57.5 & 45.4 \\
    DS-R1-Qwen-32B & 23.3 & 18.1 & 49.0 & 45.1 & 96.0 & 48.6 & 85.7 & 79.7 & 15.4 & 7.9 & 53.9 & 39.9 \\
    DS-R1-Qwen-14B & 18.0 & 13.4 & 52.0 & 51.7 & 32.0 & 16.2 & 67.3 & 63.4 & 4.5 & 1.4 & 34.8 & 29.2 \\
    DS-R1-Qwen-7B & 6.3 & 1.9 & 1.0 & 0.9 & 2.0 & 1.0 & 13.7 & 8.9 & 0.5 & 0.1 & 4.7 & 2.5 \\
    Qwen2.5-72B & 34.7 & 15.2 & 59.0 & 58.5 & 90.0 & 39.5 & 27.0 & 17.4 & 6.5 & 2.1 & 43.4 & 26.5 \\
    Qwen2.5-32B & 35.3 & 15.2 & 15.0 & 12.8 & \textbf{100.0} & 51.7 & 32.0 & 22.4 & 23.5 & 6.7 & 41.2 & 21.8 \\
    Qwen2.5-14B & 30.0 & 11.5 & 21.0 & 15.8 & 89.0 & 44.3 & 23.3 & 12.7 & 17.5 & 4.9 & 36.2 & 17.8 \\
    Qwen2.5-3B & 7.7 & 2.9 & 17.0 & 5.0 & 6.0 & 2.2 & 23.0 & 10.7 & 15.5 & 3.7 & 13.8 & 4.9 \\
    InternLM3-8b & 15.0 & 3.6 & 20.0 & 9.5 & 86.0 & 43.3 & 41.3 & 23.7 & 16.0 & 4.1 & 35.7 & 16.8 \\
    LLama3.1-8B & 23.0 & 8.0 & 9.0 & 7.8 & 0.0 & 0.0 & 28.7 & 11.4 & 15.0 & 1.8 & 15.1 & 5.8 \\
    \midrule
    Qwen2.5-7B & 11.0 & 3.1 & 40.0 & 19.8 & 67.0 & 34.0 & 26.7 & 15.2 & 3.5 & 1.0 & 29.6 & 14.6 \\
    {\model-7B-NP (Ours)} & 89.7 & 27.8 & 85.0 & 43.5 & 99.0 & 53.8 & 93.7 & 79.1 & 98.2 & 28.9 & \textbf{93.1} & 46.6 \\
    \rowcolor{lightgray}
     \cellcolor{white}& \textcolor{red}{+78.7} & \textcolor{red}{+24.7} & \textcolor{red}{+45.0} & \textcolor{red}{+23.7} & \textcolor{red}{+32.0} & \textcolor{red}{+19.8} & \textcolor{red}{+67.0} & \textcolor{red}{+63.9} & \textcolor{red}{+94.7} & \textcolor{red}{+27.9} & \textcolor{red}{+63.5} & \textcolor{red}{+32.0} \\
    \textsc{Qwen2.5-7B-SFT}(Ours) & 35.0 & 29.3 & 63.0 & 57.2 & 96.0 & 49.7 & 68.0 & 62.0 & 22.0 & 16.8 & 56.8 & 43.0 \\
    \rowcolor{lightgray}
    \cellcolor{white}& \textcolor{red}{+24.0} & \textcolor{red}{+26.2} & \textcolor{red}{+23.0} & \textcolor{red}{+37.4} & \textcolor{red}{+29.0} & \textcolor{red}{+15.7} & \textcolor{red}{+41.3} & \textcolor{red}{+46.8} & \textcolor{red}{+18.5} & \textcolor{red}{+15.8} & \textcolor{red}{+27.2} & \textcolor{red}{+28.4} \\
    {\model-7B-NP-MATH (Ours)} & 89.0 & 32.2 & 72.0 & 57.0 & \textbf{100.0} & 51.8 & 96.3 & 83.4 & \textbf{99.0} & 32.3 & 91.3 & 51.3 \\
    \rowcolor{lightgray}
    \cellcolor{white}& \textcolor{red}{+78.0} & \textcolor{red}{+29.1} & \textcolor{red}{+32.0} & \textcolor{red}{+37.2} & \textcolor{red}{+33.0} & \textcolor{red}{+17.8} & \textcolor{red}{+69.6} & \textcolor{red}{+68.2} & \textcolor{red}{+95.5} & \textcolor{red}{+31.3} & \textcolor{red}{+61.7} & \textcolor{red}{+36.7} \\
    \bottomrule
    \end{tabular}
    }
    \label{tab:indomain}
    \vspace{-1em}
\end{table*}
\begin{table*}[t]
    \caption{Performance on out-of-domain benchmarks, including both reasoning and non-reasoning tasks, demonstrates that RLVR training on \data generalizes effectively.}
    \centering
    \small
    \resizebox{\textwidth}{!}{
    \begin{tabular}{lcccccc}
    \toprule
    \multirow{2}{*}{\textbf{Model}} &
      \textbf{Logic} & \multicolumn{2}{c}{\textbf{Math}} & \textbf{Knowledge} & \textbf{Instruction} & \multirow{2}{*}{\textbf{Average}} \\
    \cmidrule(lr){2-2} \cmidrule(lr){3-4} \cmidrule(lr){5-5} \cmidrule(lr){6-6}
     & 
      \textbf{KORBench} & \textbf{Math500} & \textbf{OlpBench} & \textbf{GPQA\_diamond} & \textbf{IFEval} \\
    \midrule
    Qwen2.5-72B & 53.0 & \textbf{84.2} & \textbf{49.1} & \textbf{46.0} & \textbf{84.3} & \textbf{63.3} \\
    Qwen2.5-32B & \textbf{56.2} & 82.8 & 48.8 & 43.9 & 79.5 & 62.3 \\
    Qwen2.5-14B & 50.6 & 80.0 & 45.1 & 41.9 & 81.6 & 59.8 \\
    Qwen2.5-3B & 36.6 & 67.2 & 29.7 & 30.3 & 59.1 & 44.6 \\
    InternLM3-8B & 40.7 & 78.2 & 25.1 & 36.9 & 72.5 & 50.7 \\
    LLama3.1-8b & 44.5 & 48.6 & 17.7 & 22.7 & 69.3 & 40.6 \\
    \midrule
    Qwen2.5-7B & 42.9 & 72.4 & 30.0 & 33.3 & 73.5 & 50.4 \\
    \model-7B-NP & {44.1 (\textcolor{red}{+1.2})} &  {74.6 (\textcolor{red}{+2.2})} & 32.1 (\textcolor{red}{+2.1}) &  {37.4 (\textcolor{red}{+4.1})} &  {79.6 (\textcolor{red}{+6.1})} &  {53.6 (\textcolor{red}{+3.2})} \\
    \model-7B-MATH & {45.9 (\textcolor{red}{+3.0})} & 77.4 (\textcolor{red}{+5.0}) & 31.8 (\textcolor{red}{+1.8}) & 33.8 (\textcolor{red}{+0.5}) & 80.5 (\textcolor{red}{+7.0}) & 53.9 (\textcolor{red}{+3.5}) \\
    Qwen2.5-7B-SFT & 30.1 (\textcolor{green}{-12.8}) & 66.2 (\textcolor{green}{-6.2}) & 21.4 (\textcolor{green}{-8.6}) & 18.7 (\textcolor{green}{-14.6}) & 55.8 (\textcolor{green}{-17.7}) & 38.4 (\textcolor{green}{-12.0}) \\
    \model-7B-NP-MATH & 46.6 (\textcolor{red}{+3.7}) & 75.0 (\textcolor{red}{+2.6}) & 31.2 (\textcolor{red}{+1.2}) & 38.9 (\textcolor{red}{+5.6}) & {79.3 (\textcolor{red}{+5.8})} & 54.2 (\textcolor{red}{+3.8}) \\
    \bottomrule
    \end{tabular}
    }
    \label{tab:ood}
    \vspace{-0.5em}
\end{table*}

\paragraph{RLVR vs. SFT.} To isolate the contribution of RLVR, we compare \model against a SFT baseline (\textbf{Qwen2.5-7B-SFT}) trained on the same optimal solutions. While SFT improves in-domain performance (56.8\% SR) over the base model, it lags significantly behind \model (93.1\% SR). Crucially, SFT suffers from catastrophic out-of-domain degradation (dropping 12.0 points to an average score of 38.4), whereas \model improves generalization (53.6). We attribute this disparity to SFT's \textbf{lack of exploration}: by strictly mimicking optimal solutions, SFT overfits to specific solution patterns without mastering the underlying construction logic. In contrast, RLVR forces the model to explore diverse trajectories—including suboptimal and invalid attempts—enabling the acquisition of robust, transferable constraint satisfaction strategies rather than rote memorization.

\subsection{Ablation Studies}



\begin{table*}[t]
    \caption{
Comparison of curriculum learning performance across different RL algorithms from \data during RLVR training. Additionally, the we compares the impact of fine-grained versus coarse reward designs.
}
    \centering
    \small
    \resizebox{\textwidth}{!}{
    \begin{tabular}{lccccccccc}
    \toprule
    \multirow{2}{*}{\textbf{Data Proportion}} & \textbf{CL} &
      \multicolumn{2}{c}{\textbf{In Domain}} &
      \multicolumn{6}{c}{\textbf{Out of Domain}} \\
    \cmidrule(lr){3-4} \cmidrule(lr){5-10}
     & & 
      \textbf{SR} & \textbf{QR} &
      \textbf{KB} & \textbf{Math500} & \textbf{OB} & \textbf{GPQA} & \textbf{IF} & \textbf{Avg} \\
    \midrule
    Qwen2.5-7B (Base) & \ding{55} & 29.6 & 14.6 & 42.9 & 72.4 & 30.0 & 33.3 & 73.5 & 50.4 \\
    Qwen2.5-7B (binary)-GRPO &\ding{51} & 86.6 & 17.8 & 43.4 & 73.0 & 30.5 & 30.3 & 78.3 & 51.1\\
    Qwen2.5-7B (w/o GT)-GRPO &\ding{51} & 76.2 & 25.6 & 43.8 & 73.2 & 30.5 & 30.8 & 78.3 & 51.3 \\
    \midrule
    \multirow{2}{*}{\model-7B-GPG~\cite{gpg}} & \ding{55} & 85.8 & 32.6 & 43.8 & 74.2 & 30.6 & 30.8 & 78.0 & 51.5 \\
                                & \ding{51} & 88.6 & 32.7 & \textbf{44.9} & 74.2 & 30.9 & \textbf{38.9} & 78.6 & 53.5 \\
    \midrule
    \multirow{2}{*}{\model-7B-GSPO~\cite{gspo}} & \ding{55} & 78.2 & 36.6 & 40.8 & \textbf{75.6} & 31.8 & 36.4 & 78.2 & 52.6 \\
                                & \ding{51} & 83.4 & 42.1 & 44.4 & 75.0 & 31.1 & 36.9 & 77.2 & 52.9 \\
    \midrule
    \multirow{2}{*}{\model-7B-GRPO~\cite{guo2025deepseek}} &\ding{55} & 80.8 & 37.4 & 43.8 & 75.2 & 31.7 & 34.3 & 77.9 & 52.6  \\
    &\ding{51} & \textbf{93.1} & \textbf{46.6} & 44.1 & 74.6 & \textbf{32.1} & 37.4 & \textbf{79.6} & \textbf{53.6} \\
    \bottomrule
    \end{tabular}
    }
    \label{tab:reward}
    \vspace{-1em}
\end{table*}
\subsubsection{Impact of Quality-Aware Rewards}
Table~\ref{tab:reward} demonstrates the critical impact of reward design. Compared with binary rewards (Qwen2.5-7B (binary)-GRPO) and coarse rewards (Qwen2.5-7B (w/o GT)-GRPO), our fine-grained quality-aware rewards yield substantial improvements in both solution quality and success rate. Notably, reward schemes that focus solely on feasibility and assign a fixed reward of 1.0 tend to induce reward hacking in the early training stage, since finding a feasible solution is relatively easy for LLMs. These results confirm that explicit optimality feedback not only guides the model toward better solutions, but also strengthens its underlying mastery of feasibility constraints.




\paragraph{Curriculum Design and Strategy.} 
As shown in Table~\ref{tab:single_task}, we first analyze the impact of instance difficulty distribution by varying the ratio of Easy, Medium, and Hard examples. Results indicate that an \textit{easy-heavy} configuration (5:4:1) yields the best in-domain performance (100\% SR) and strongest out-of-distribution (OOD) transfer. This supports the hypothesis that mastering foundational skills on simpler tasks is essential for both advanced optimization and generalizable reasoning.

We evaluate \textit{Curriculum Replay} against single-pass baselines, analyzing both overall performance (Figure~\ref{fig:comp}) and training reward dynamics (Figure~\ref{fig:rewardplot}). Quantitatively, Replay breaks the performance ceiling of linear strategies, achieving 93.1\% SR and a +15.1 boost in QR. The reward trajectories reveal the underlying mechanism: while the Linear strategy (purple) suffers from early saturation, Replay (green) exhibits higher volatility—a natural byproduct of cyclically revisiting difficulty levels. This dynamic re-calibration allows the policy to escape local optima, preventing stagnation and mitigating catastrophic forgetting to sustain a continuous upward trajectory.

\paragraph{Performance on Large and Reasoning LLMs.}
Table~\ref{tab:large_reasoning} shows two key findings. First, FORGE data scales well to larger base models. On \texttt{Qwen2.5-14B-Instruct}, our method markedly improves in-domain optimization performance, increasing SR from 36.2 to 78.6 and QR from 17.8 to 48.4, outperforming the GRPO Shuffle method. These gains also transfer to out-of-domain reasoning benchmarks, where the average score rises from 59.8 to 63.1. This suggests that FORGE data not only improves optimization on \bench, but also unlocks stronger latent reasoning ability in larger foundation models.

Second, general reasoning ability does not directly yield strong optimization performance. Although \texttt{DeepSeek-R1-Distill-7B} is a strong reasoning model, its base version performs poorly on \bench, with only 4.7 SR and 2.5 QR. After RLVR training on Forge Data, our method improves the model to 32.3 SR and 19.2 QR, showing that it effectively instills optimization-specific reasoning behaviors. These gains also extend beyond the training domain, improving the average out-of-domain score from 53.8 to 58.4. This supports our claim that optimization reasoning learned from NP-hard problem can transfer to broader reasoning tasks.
\begin{table*}[t]
\caption{
    Performance of RLVR training on Forge Data for large language models and reasoning models.
}
    \centering
    \small
    \resizebox{\textwidth}{!}{
    \begin{tabular}{lcccccccc}
    \toprule
    \multirow{2}{*}{\textbf{Method}} &
    \multicolumn{2}{c}{\textbf{In Domain}} &
    \multicolumn{6}{c}{\textbf{Out of Domain}} \\
    \cmidrule(lr){2-3} \cmidrule(lr){4-9}
    & \textbf{SR} & \textbf{QR}
    & \textbf{KB} & \textbf{Math500} & \textbf{OB} & \textbf{GPQA} & \textbf{IF} & \textbf{Avg} \\
    \midrule
    Qwen2.5-14B-Instruct (Base) & 36.2 & 17.8 & 50.6 & 80.0 & 45.1 & 41.9 & 81.6 & 59.8 \\
    Qwen2.5-14B-Instruct (GRPO Shuffle) & 77.2 & 46.4 & 54.8 & 81.8 & 46.3 & 47.1 & 82.1 & 62.4 \\
    Qwen2.5-14B-Instruct (Froge) & \textbf{78.6} & \textbf{48.4} & \textbf{55.6} & \textbf{82.4} & \textbf{47.2} & \textbf{47.3} & \textbf{83.1} & \textbf{63.1} \\
    \midrule
    DeepSeek-R1-Distill-7B (Base) & 4.7 & 2.5 & 51.5 & 72.8 & 40.1 & 32.8 & 71.7 & 53.8 \\
    DeepSeek-R1-Distill-7B (GRPO Shuffle) & 29.2 & 17.6 & 51.9 & 78.4 & 35.8 & \textbf{45.2} & 75.1 & 57.3 \\
    DeepSeek-R1-Distill-7B (Froge) & \textbf{32.3} & \textbf{19.2} & \textbf{52.7} & \textbf{81.6} & \textbf{46.2} & 36.4 & \textbf{75.3} & \textbf{58.4} \\
    \bottomrule
    \end{tabular}
    }
    \label{tab:large_reasoning}
    \vspace{-1em}
\end{table*}
\section{Related Work}
\textbf{Reinforcement Learning with Verifiable Rewards (RLVR).}  
As reinforcement learning (RL) becomes an increasingly important tool for enhancing the reasoning capabilities of LLMs, Reinforcement Learning with Verifiable Rewards (RLVR) has emerged as a compelling alternative to Reinforcement Learning with Human Feedback (RLHF). Unlike RLHF, which relies on pretrained reward models and subjective human annotations, RLVR utilizes objective, automatically verifiable outcomes to provide reliable supervision~\cite{seed2025seed,guo2025deepseek,team2025kimi,timelymachine,contextRL}. Recent models exemplify this paradigm shift: DeepSeek-R1~\cite{guo2025deepseek} improves long-chain reasoning and self-verification through RLVR, while Kimi K1.5~\cite{team2025kimi} achieves strong performance with long-context training and streamlined policy optimization—without depending on complex value models.
The ecosystem supporting RLVR is rapidly maturing. High-quality math corpora with verifiable solutions~\cite{he2025deepmath,albalak2025big}, structured coding corpora with graded difficulty and reward pipelines~\cite{liu2025code,xu2025kodcode}, and procedurally generated puzzle-style datasets with algorithmic verification~\cite{xie2025logic,enigmata,internbootcamp} are now available. Notably, NP problems are inherently verifiable and offer controllable difficulty settings~\cite{nphardeval,yang2025nppc}, making them well-suited for RLVR-based training. However, most prior RLVR efforts have focused on math, coding, logic, or puzzles, leaving the broader class of NP-hard problems underexplored.


\noindent
\textbf{Optimization Reasoning with LLMs.}  
Various benchmarks have been proposed to evaluate LLMs' reasoning capabilities across different domains, including mathematical~\cite{glazer2024frontiermath}, logical~\cite{xie2025logic}, puzzle~\cite{enigmata}, and programming reasoning~\cite{xu2025icpc}. These tasks typically involve binary answer validation (e.g., True or False), which primarily assess deductive or symbolic reasoning. In contrast, optimization reasoning presents a fundamentally different challenge: it requires models to generate not only feasible solutions but also solutions that are as optimal as possible. 
Previous work on NP tasks has faced challenges in trainability~\cite{nphardeval,yang2025nppc}. Despite its significance, optimization reasoning has been underexplored in RLVR-based LLM training. Our work addresses this gap by focusing on NP-class problems. We propose \data, a unified framework for data generation, optimal solution annotation, RLVR training, and evaluation, which empowers LLMs with optimization reasoning capabilities.
\section{Conclusion}
We presented \data, a comprehensive framework that leverages scalable generation and heuristic-guided rewards to train LLMs on NP-hard optimization problems. Through this infrastructure and the accompanying \bench benchmark, we developed \model, which significantly outperforms GPT-4o in optimization reasoning. Our experiments validate the effectiveness of quality-aware reward design and multi-stage curriculum learning. Furthermore, we reveal a critical insight: task diversity in RLVR training is a key driver of out-of-domain generalization. 
We hope this work lays a foundation for future research on integrating LLMs with optimization-based reasoning, offering new insights to the community and advancing the frontier of LLM reasoning capabilities.

\section*{Limitations}

Due to computational resource constraints, we have conducted experiments using the Qwen2.5-7B-Instruct-1M model. Larger models, such as those with 14B or 32B parameters, have not been trained, and the performance of these more powerful models, starting from a bigger model size, remains unexplored. Additionally, designing individual NP tasks for RLVR training requires meticulous attention to various aspects, including problem definition, validation script development, heuristic algorithm design, and difficulty level calibration. As a result, we have currently designed 10 tasks. Scaling to larger models and incorporating additional tasks remain avenues for future exploration.




\bibliography{custom}

@article{borazjanizadeh2024navigating,
  title={Navigating the Labyrinth: Evaluating and Enhancing LLMs' Ability to Reason About Search Problems},
  author={Borazjanizadeh, Nasim and Herzig, Roei and Darrell, Trevor and Feris, Rogerio and Karlinsky, Leonid},
  journal={arXiv preprint arXiv:2406.12172},
  year={2024}
}

@article{lin2025zebralogic,
  title={ZebraLogic: On the Scaling Limits of LLMs for Logical Reasoning},
  author={Lin, Bill Yuchen and Bras, Ronan Le and Richardson, Kyle and Sabharwal, Ashish and Poovendran, Radha and Clark, Peter and Choi, Yejin},
  journal={arXiv preprint arXiv:2502.01100},
  year={2025}
}

@article{wu2025phd,
  title={Phd knowledge not required: A reasoning challenge for large language models},
  author={Wu, Zixuan and Lucchetti, Francesca and Boruch-Gruszecki, Aleksander and Zhao, Jingmiao and Anderson, Carolyn Jane and Biswas, Joydeep and Cassano, Federico and Feldman, Molly Q and Guha, Arjun},
  journal={arXiv preprint arXiv:2502.01584},
  year={2025}
}

@article{ma2024kor,
  title={Kor-bench: Benchmarking language models on knowledge-orthogonal reasoning tasks},
  author={Ma, Kaijing and Du, Xinrun and Wang, Yunran and Zhang, Haoran and Wen, Zhoufutu and Qu, Xingwei and Yang, Jian and Liu, Jiaheng and Liu, Minghao and Yue, Xiang and others},
  journal={arXiv preprint arXiv:2410.06526},
  year={2024}
}

@article{albalak2025big,
  title={Big-Math: A Large-Scale, High-Quality Math Dataset for Reinforcement Learning in Language Models},
  author={Albalak, Alon and Phung, Duy and Lile, Nathan and Rafailov, Rafael and Gandhi, Kanishk and Castricato, Louis and Singh, Anikait and Blagden, Chase and Xiang, Violet and Mahan, Dakota and others},
  journal={arXiv preprint arXiv:2502.17387},
  year={2025}
}

@article{he2025deepmath,
  title={DeepMath-103K: A Large-Scale, Challenging, Decontaminated, and Verifiable Mathematical Dataset for Advancing Reasoning},
  author={He, Zhiwei and Liang, Tian and Xu, Jiahao and Liu, Qiuzhi and Chen, Xingyu and Wang, Yue and Song, Linfeng and Yu, Dian and Liang, Zhenwen and Wang, Wenxuan and others},
  journal={arXiv preprint arXiv:2504.11456},
  year={2025}
}

@article{xu2025kodcode,
  title={Kodcode: A diverse, challenging, and verifiable synthetic dataset for coding},
  author={Xu, Zhangchen and Liu, Yang and Yin, Yueqin and Zhou, Mingyuan and Poovendran, Radha},
  journal={arXiv preprint arXiv:2503.02951},
  year={2025}
}

@article{xie2025logic,
  title={Logic-rl: Unleashing llm reasoning with rule-based reinforcement learning},
  author={Xie, Tian and Gao, Zitian and Ren, Qingnan and Luo, Haoming and Hong, Yuqian and Dai, Bryan and Zhou, Joey and Qiu, Kai and Wu, Zhirong and Luo, Chong},
  journal={arXiv preprint arXiv:2502.14768},
  year={2025}
}

@article{team2025kimi,
  title={Kimi k1. 5: Scaling reinforcement learning with llms},
  author={Team, Kimi and Du, Angang and Gao, Bofei and Xing, Bowei and Jiang, Changjiu and Chen, Cheng and Li, Cheng and Xiao, Chenjun and Du, Chenzhuang and Liao, Chonghua and others},
  journal={arXiv preprint arXiv:2501.12599},
  year={2025}
}

@misc{o1,
  title={Learning to reason with LLMs},
  author={OpenAI},
  year={2024},
  url = {https://openai.com/index/learning-to-reason-with-llms/}
}

@misc{claude,
  title={Claude 3.7 Sonnet and Claude Code},
  author={Anthropic},
  year={2025},
  url = {https://www.anthropic.com/news/claude-3-7-sonnet}
}

@misc{gemini,
  title={Gemini 2.5: Our most intelligent AI model},
  author={Google},
  year={2025},
  url = {https://blog.google/technology/google-deepmind/gemini-model-thinking-updates-march-2025/#gemini-2-5-thinking}
}

@article{guo2025deepseek,
  title={Deepseek-r1: Incentivizing reasoning capability in llms via reinforcement learning},
  author={Guo, Daya and Yang, Dejian and Zhang, Haowei and Song, Junxiao and Zhang, Ruoyu and Xu, Runxin and Zhu, Qihao and Ma, Shirong and Wang, Peiyi and Bi, Xiao and others},
  journal={arXiv preprint arXiv:2501.12948},
  year={2025}
}

@article{liu2025code,
  title={Code-r1: Reproducing r1 for code with reliable rewards},
  author={Liu, Jiawei and Zhang, Lingming},
  journal={arXiv preprint arXiv:2503.18470},
  year={2025}
}

@article{seed2025seed,
  title={Seed-thinking-v1. 5: Advancing superb reasoning models with reinforcement learning},
  author={Seed, ByteDance and Yuan, Yufeng and Yue, Yu and Wang, Mingxuan and Zuo, Xiaochen and Chen, Jiaze and Yan, Lin and Xu, Wenyuan and Zhang, Chi and Liu, Xin and others},
  journal={arXiv preprint arXiv:2504.13914},
  year={2025}
}

@misc{enigmata,
      title={Enigmata: Scaling Logical Reasoning in Large Language Models with Synthetic Verifiable Puzzles}, 
      author={Jiangjie Chen and Qianyu He and Siyu Yuan and Aili Chen and Zhicheng Cai and Weinan Dai and Hongli Yu and Qiying Yu and Xuefeng Li and Jiaze Chen and Hao Zhou and Mingxuan Wang},
      year={2025},
      eprint={2505.19914},
      archivePrefix={arXiv},
      primaryClass={cs.CL},
      url={https://arxiv.org/abs/2505.19914}, 
}

@inproceedings{nphardeval,
  title={NPHardEval: Dynamic Benchmark on Reasoning Ability of Large Language Models via Complexity Classes},
  author={Fan, Lizhou and Hua, Wenyue and Li, Lingyao and Ling, Haoyang and Zhang, Yongfeng},
  booktitle={62nd Annual Meeting of the Association for Computational Linguistics, ACL 2024},
  pages={4092--4114},
  year={2024},
  organization={Association for Computational Linguistics (ACL)}
}

@article{yang2025nppc,
  title={Nondeterministic Polynomial-time Problem Challenge: An Ever-Scaling Reasoning Benchmark for LLMs},
  author={Yang, Chang and Wang, Ruiyu and Jiang, Junzhe and Jiang, Qi and Zhang, Qinggang and Deng, Yanchen and Li, Shuxin and Hu, Shuyue and Li, Bo and Pokorny, Florian T and others},
  journal={arXiv preprint arXiv:2504.11239},
  year={2025}
}

@article{glazer2024frontiermath,
  title={Frontiermath: A benchmark for evaluating advanced mathematical reasoning in ai},
  author={Glazer, Elliot and Erdil, Ege and Besiroglu, Tamay and Chicharro, Diego and Chen, Evan and Gunning, Alex and Olsson, Caroline Falkman and Denain, Jean-Stanislas and Ho, Anson and Santos, Emily de Oliveira and others},
  journal={arXiv preprint arXiv:2411.04872},
  year={2024}
}

@article{xu2025icpc,
  title={ICPC-Eval: Probing the Frontiers of LLM Reasoning with Competitive Programming Contests},
  author={Xu, Shiyi and Hu, Yiwen and Min, Yingqian and Chen, Zhipeng and Zhao, Wayne Xin and Wen, Ji-Rong},
  journal={arXiv preprint arXiv:2506.04894},
  year={2025}
}

@misc{yang2025qwen251mtechnicalreport,
      title={Qwen2.5-1M Technical Report}, 
      author={An Yang and Bowen Yu and Chengyuan Li and Dayiheng Liu and Fei Huang and Haoyan Huang and Jiandong Jiang and Jianhong Tu and Jianwei Zhang and Jingren Zhou and Junyang Lin and Kai Dang and Kexin Yang and Le Yu and Mei Li and Minmin Sun and Qin Zhu and Rui Men and Tao He and Weijia Xu and Wenbiao Yin and Wenyuan Yu and Xiafei Qiu and Xingzhang Ren and Xinlong Yang and Yong Li and Zhiying Xu and Zipeng Zhang},
      year={2025},
      eprint={2501.15383},
      archivePrefix={arXiv},
      primaryClass={cs.CL},
      url={https://arxiv.org/abs/2501.15383}, 
}

@misc{internbootcamp,
      title={InternBootcamp Technical Report: Boosting LLM Reasoning with Verifiable Task Scaling}, 
      author={Peiji Li and Jiasheng Ye and Yongkang Chen and Yichuan Ma and Zijie Yu and Kedi Chen and Ganqu Cui and Haozhan Li and Jiacheng Chen and Chengqi Lyu and Wenwei Zhang and Linyang Li and Qipeng Guo and Dahua Lin and Bowen Zhou and Kai Chen},
      year={2025},
      eprint={2508.08636},
      archivePrefix={arXiv},
      primaryClass={cs.CL},
      url={https://arxiv.org/abs/2508.08636}, 
}

@misc{2023opencompass,
    title={OpenCompass: A Universal Evaluation Platform for Foundation Models},
    author={OpenCompass Contributors},
    howpublished = {\url{https://github.com/open-compass/opencompass}},
    year={2023}
}

@misc{yu2025dapo17k,
      title={DAPO: An Open-Source LLM Reinforcement Learning System at Scale}, 
      author={Qiying Yu and Zheng Zhang and Ruofei Zhu and Yufeng Yuan and Xiaochen Zuo and Yu Yue and Weinan Dai and Tiantian Fan and Gaohong Liu and Lingjun Liu and Xin Liu and Haibin Lin and Zhiqi Lin and Bole Ma and Guangming Sheng and Yuxuan Tong and Chi Zhang and Mofan Zhang and Wang Zhang and Hang Zhu and Jinhua Zhu and Jiaze Chen and Jiangjie Chen and Chengyi Wang and Hongli Yu and Yuxuan Song and Xiangpeng Wei and Hao Zhou and Jingjing Liu and Wei-Ying Ma and Ya-Qin Zhang and Lin Yan and Mu Qiao and Yonghui Wu and Mingxuan Wang},
      year={2025},
      eprint={2503.14476},
      archivePrefix={arXiv},
      primaryClass={cs.LG},
      url={https://arxiv.org/abs/2503.14476}, 
}

@misc{gspo,
      title={Group Sequence Policy Optimization}, 
      author={Chujie Zheng and Shixuan Liu and Mingze Li and Xiong-Hui Chen and Bowen Yu and Chang Gao and Kai Dang and Yuqiong Liu and Rui Men and An Yang and Jingren Zhou and Junyang Lin},
      year={2025},
      eprint={2507.18071},
      archivePrefix={arXiv},
      primaryClass={cs.LG},
      url={https://arxiv.org/abs/2507.18071}, 
}

@misc{gpg,
      title={GPG: A Simple and Strong Reinforcement Learning Baseline for Model Reasoning}, 
      author={Xiangxiang Chu and Hailang Huang and Xiao Zhang and Fei Wei and Yong Wang},
      year={2025},
      eprint={2504.02546},
      archivePrefix={arXiv},
      primaryClass={cs.LG},
      url={https://arxiv.org/abs/2504.02546}, 
}

@misc{timelymachine,
      title={Timely Machine: Awareness of Time Makes Test-Time Scaling Agentic}, 
      author={Yichuan Ma and Linyang Li and Yongkang chen and Peiji Li and Xiaozhe Li and Qipeng Guo and Dahua Lin and Kai Chen},
      year={2026},
      eprint={2601.16486},
      archivePrefix={arXiv},
      primaryClass={cs.CL},
      url={https://arxiv.org/abs/2601.16486}, 
}

@misc{contextRL,
      title={Escaping the Context Bottleneck: Active Context Curation for LLM Agents via Reinforcement Learning}, 
      author={Xiaozhe Li and Tianyi Lyu and Yizhao Yang and Liang Shan and Siyi Yang and Ligao Zhang and Zhuoyi Huang and Qingwen Liu and Yang Li},
      year={2026},
      eprint={2604.11462},
      archivePrefix={arXiv},
      primaryClass={cs.AI},
      url={https://arxiv.org/abs/2604.11462}, 
}

\appendix
\newpage
\section{Appendix}

\subsection{Use of Large Language Models}
Large Language Models are used for grammar check and polishing in this paper.
\subsection{Ethics Statement.}
All datasets used in this study are obtained from public sources and are freely available for academic research. Therefore, we do not anticipate that the data used in this work pose any significant privacy risks.
\subsection{Detailed Training Setting}
\label{apd:setting}
\subsubsection{RLVR Training Setting}
The training experiment utilizes the verl framework, employing the GRPO algorithm for fine-tuning the Qwen2.5-7B-Instruct-1M model. Training is performed on 8 A800 GPUs with a batch size of 16 for both training and validation. The maximum prompt length is set to 20,000 tokens, and the response length is capped at 4,096 tokens. Key hyperparameters include a learning rate of \(4 \times 10^{-7}\), with a mini-batch size of 16 and a micro-batch size of 16 for PPO updates. KL loss regularization, with a coefficient of \(0.001\) (\(\text{KL}_\text{coef}\)), is applied to stabilize training.

\subsubsection{SFT Training Implementation}

\textbf{Data Curation.} 
We derived our SFT dataset through knowledge distillation from the \texttt{Qwen3-235B-Thinking} teacher model. To ensure data quality, we applied Rejection Sampling Fine-Tuning (RFT), filtering generated trajectories based on their Quality Ratio (QR). Specifically, we selected solutions with $QR > 0.5$, indicating that the solution quality is at least 50\% of the heuristic optimal baseline. This process yielded a dataset of 10,000 samples, evenly distributed with 1,000 instances across each of the 10 NP tasks.

\noindent\textbf{Training Details.} 
We employed the ms-Swift framework for full-parameter fine-tuning, leveraging DeepSpeed ZeRO-2 and \texttt{bfloat16} precision to optimize computational efficiency. The model was trained for 2 epochs with a learning rate of $1\times10^{-5}$ and a warmup ratio of 0.1. To accommodate the extensive context required for NP-hard reasoning, we set the maximum sequence length to 20,480 tokens. For memory management, we utilized a per-device batch size of 2 combined with 16 gradient accumulation steps.

\subsection{Experiment Results}
We conduct comprehensive experiments to answer three key questions: (1) Can LLMs learn to generate near-optimal solutions through quality-aware RLVR? (2) Does optimization training transfer to general reasoning? (3) What training strategies are most effective? Our experiments demonstrate that \model achieves state-of-the-art optimization performance (93.1\% SR, 46.6\% AR), transfers to diverse reasoning tasks (+3.2 points OOD), and reveals that task diversity and quality-aware rewards are critical for effective learning.
\subsubsection{Curriculum Learning and Data Proportion}
\begin{table*}[t]
    \caption{Comparison of different data proportions for easy (E), medium (M), and hard (H) from \data during RLVR, as well as curriculum learning (CL) strategies. Training on single TSP problem.}
    \centering
    \small
    \resizebox{\textwidth}{!}{
    \begin{tabular}{lccccccccc}
    \toprule
    \multirow{2}{*}{\textbf{Data Proportion}} & \textbf{CL} &
      \multicolumn{2}{c}{\textbf{In Domain}} &
      \multicolumn{6}{c}{\textbf{Out of Domain}} \\
    \cmidrule(lr){3-4} \cmidrule(lr){5-10}
     & & 
      \textbf{SR} & \textbf{AR} &
      \textbf{KB} & \textbf{Math500} & \textbf{OB} & \textbf{GPQA} & \textbf{IF} & \textbf{Avg} \\
    \midrule
    Qwen2.5-7B (baseline) & \ding{55} & 4.0 & 1.9 & 42.9 & 72.4 & 30.0 & 33.3 & 73.5 & 50.4 \\
    \midrule
    \multirow{2}{*}{E:M:H=1:4:5} & \ding{55} & 99.0 & 28.1 & 42.9 & 73.8 & \textbf{31.1} & 35.4 & 77.9 & 52.2 \\
                                & \ding{51} & 98.0 & 28.2 & 43.4 & 73.6 & 30.8 & \textbf{37.9} & 78.5 & 52.8 \\
    \midrule
    \multirow{2}{*}{E:M:H=1:1:1} & \ding{55} & 97.0 & 26.9 & 42.3 & 74.2 & 29.6 & 32.3 & 78.8 & 51.5 \\
                                & \ding{51} & 98.0 & 27.1 & \textbf{44.6} & \textbf{74.8} & 29.4 & 32.8 & \textbf{79.3} & 52.2 \\
    \midrule
    \multirow{2}{*}{E:M:H=5:4:1} &\ding{55} & \textbf{100.0} & 28.2 & 44.2 & 74.4 & 30.4 & 31.3 & 78.5 & 51.8  \\
    &\ding{51} & \textbf{100.0} & \textbf{29.0} & 44.3 & 74.4 & \textbf{31.1} & 35.9 & 78.6 & \textbf{52.9} \\
    \bottomrule
    \end{tabular}
    }
    \label{tab:single_task}
\end{table*}
As shown in Table~\ref{tab:single_task}, we investigate the impact of different data proportions—easy (E), medium (M), and hard (H)—on RLVR training, along with the role of curriculum learning (CL) strategies. The experiments focus on the TSP problem to better summarize the rules. We compare several configurations, including the baseline model and a variant without \heur to provide accurate ground-truth (GT) signals in the reward signal, as well as different data ratios (E:M:H).
In terms of in-domain performance, all RLVR configurations show improvements. Even the \texttt{Qwen2.5-7B (w/o GT)} model achieves noticeable gains, with SR rising from 4.0 to 90.0 and AR from 1.9 to 25.6. The introduction of curriculum learning (CL) results in further gains across all data proportions. The best performance is achieved with the \textit{E:M:H=5:4:1} configuration, which achieves an average AR of 29.0, outperforming other configurations.
For out-of-domain tasks, particularly in the \textit{Math500} and \textit{GPQA} benchmarks, the \texttt{E:M:H=5:4:1} configuration demonstrates superior generalization with an OOD average of 52.9. The inclusion of curriculum learning stabilizes performance and enhances the model's ability to generalize across both reasoning-heavy and instruction-following tasks.

Overall, these experiments highlight the importance of data proportion in RLVR, particularly the need for a larger proportion of easy tasks to build a strong foundation before tackling more complex problems. Curriculum learning further enhances this process, improving both in-domain and out-of-domain generalization capabilities.

\begin{table*}[t]
    \centering
    \small
    \begin{tabular}{lcccccc}
    \toprule
    \multirow{2}{*}{\textbf{Task Number}} &
      \textbf{Logic} & \multicolumn{2}{c}{\textbf{Math}} & \textbf{Knowledge} & \textbf{Instruction} & \multirow{2}{*}{\textbf{Average}} \\
    \cmidrule(lr){2-2} \cmidrule(lr){3-4} \cmidrule(lr){5-5} \cmidrule(lr){6-6}
     & 
      \textbf{KORBench} & \textbf{Math500} & \textbf{OlpBench} & \textbf{GPQA} & \textbf{IFEval} \\
    \midrule
    Qwen2.5-7B & 42.9 & 72.4 & 30.0 & 33.3 & 73.5 & 50.4 \\
    +3 Tasks & 43.8 (\textcolor{red}{+0.9}) & 74.8 (\textcolor{red}{+2.4}) & 30.5 (\textcolor{red}{+0.5}) & 37.4 (\textcolor{red}{+4.1}) & 78.5 (\textcolor{red}{+5.0}) & 53.0 (\textcolor{red}{+2.6}) \\
    +5 Tasks & 44.2 (\textcolor{red}{+1.3}) & 73.2 (\textcolor{red}{+0.8}) & 31.9 (\textcolor{red}{+1.9}) & 37.4 (\textcolor{red}{+4.1}) & 78.0 (\textcolor{red}{+4.5}) & 52.9 (\textcolor{red}{+2.5}) \\
    +7 Tasks & 44.2 (\textcolor{red}{+1.3}) & 76.2 (\textcolor{red}{+3.8}) & 30.4 (\textcolor{red}{+0.4}) & 35.9 (\textcolor{red}{+2.6}) & 78.8 (\textcolor{red}{+5.3}) & 53.1 (\textcolor{red}{+2.7}) \\
    +ALL Tasks & 44.1 (\textcolor{red}{+1.2}) & 74.6 (\textcolor{red}{+2.2}) & 32.1 (\textcolor{red}{+2.1}) & 37.4 (\textcolor{red}{+4.1}) & 79.6 (\textcolor{red}{+6.1}) & 53.6 (\textcolor{red}{+3.2}) \\
    \bottomrule
    \end{tabular}
    \vspace{-5pt}
    \caption{Performance on out-of-domain benchmarks, with increasing task scale from \data during RLVR training.}
    \label{tab:tsk_abl}
\end{table*}

\subsubsection{Task Diversity Drives Generalization}

Table~\ref{tab:ood} and~\ref{tab:tsk_abl} reveals a striking finding: \textbf{task diversity drives generalization more than data quantity}. Training on 10 diverse tasks achieves 53.6 OOD score, outperforming training on 3 tasks (53.0 OOD score)—despite having 3.3× less data per task. This pattern holds across all OOD benchmarks, with largest gains on mathematics (+3.8\% for 7 tasks vs. 3 tasks) and instruction-following (+6.1\% for all tasks vs. baseline).

This finding has important implications for RLVR scaling: rather than collecting massive amounts of data for few tasks, practitioners should prioritize task diversity. The results suggest that exposure to varied problem structures and constraint types develops more robust and transferable reasoning capabilities than extensive practice on narrow task distributions.

\subsubsection{Multi-Stage RL Recipe}
\begin{figure*}[ht]
    \centering
    \includegraphics[width=1\linewidth]
    {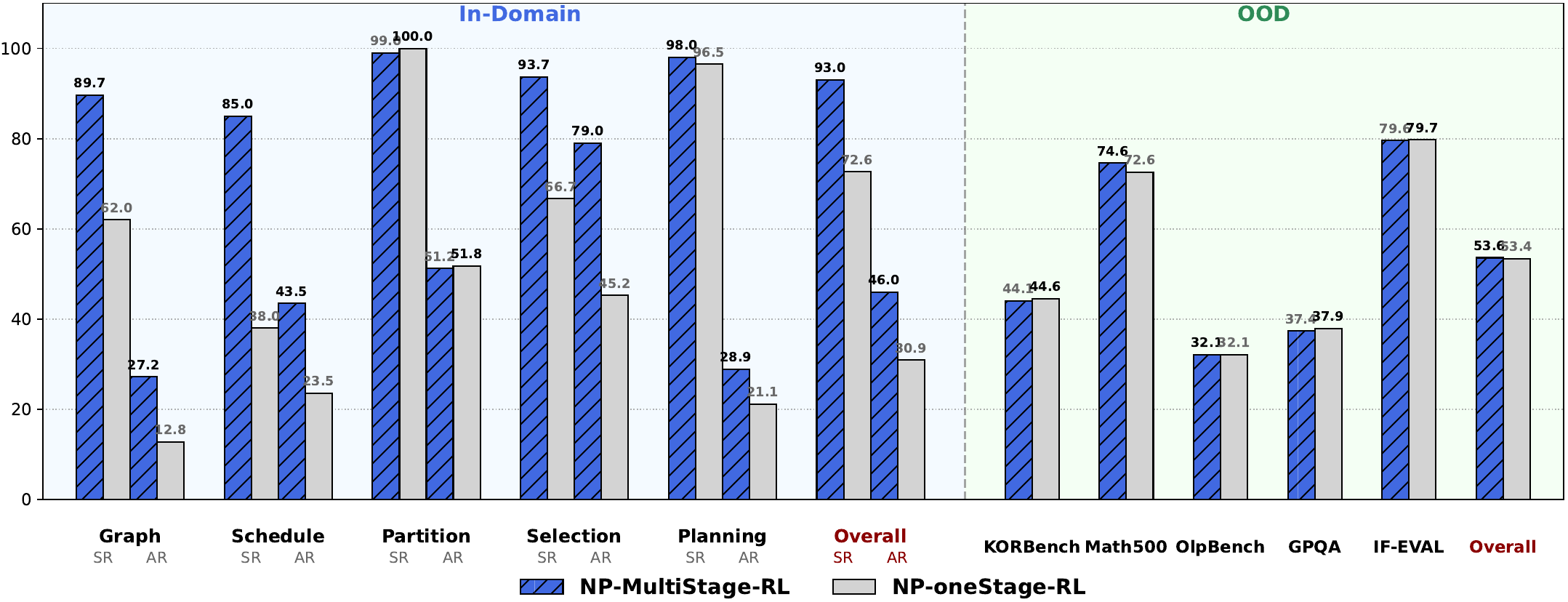}
    \caption{Comparison of RL training strategies during multi-task training, with performance evaluated on both in-domain and out-of-domain benchmarks.}
    \label{fig:comp}
\end{figure*}
\begin{figure*}[ht]
    \centering
    \includegraphics[width=\linewidth]
    {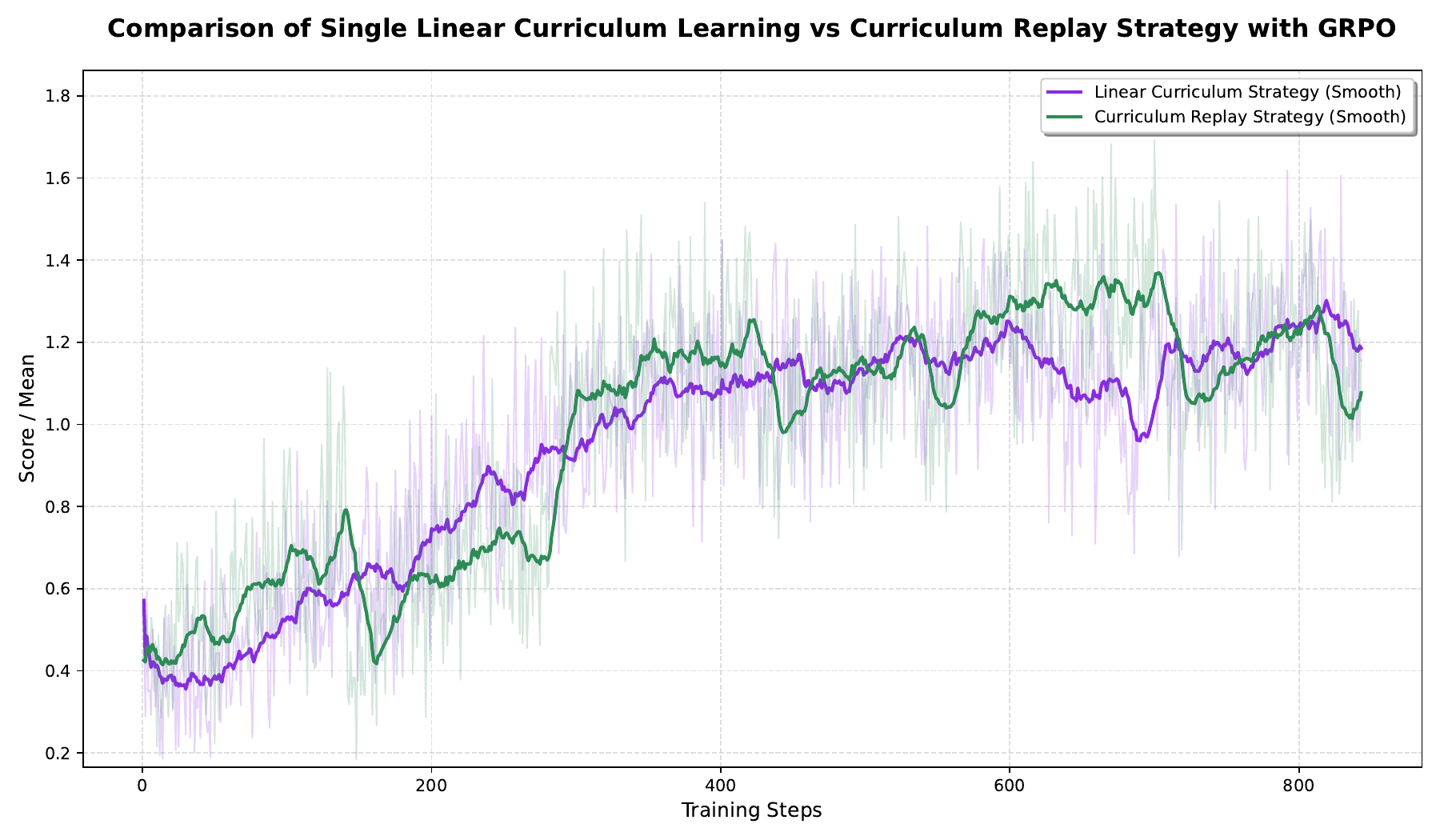}
    \caption{
    Comparison between single linear curriculum learning and curriculum replay strategy under GRPO training. The curriculum replay approach cyclically revisits samples across difficulty levels, whereas the linear strategy performs a single-pass curriculum in increasing order of difficulty.
    }
    \label{fig:rewardplot}
\end{figure*}

As illustrated in Figure~\ref{fig:comp} and Figure~\ref{fig:rewardplot}, we analyze the efficacy of the \textbf{MultiStage-RL} strategy compared to a single-pass OneStage baseline. The results reveal that iterative training is decisive for mastering complex optimization constraints. In-domain, MultiStage-RL achieves a substantial performance leap, raising the overall Success Rate (SR) from 72.6\% to 93.0\%. The gains are most pronounced in structurally complex tasks—such as \textit{Graph} (+27.7\%) and \textit{Selection} (+37.1\%)—suggesting that a linear single epoch is insufficient for the model to internalize the intricate feasibility rules of these domains.

Regarding out-of-distribution (OOD) generalization, MultiStage-RL maintains and slightly improves performance (53.6 vs. 53.4 overall). Crucially, this demonstrates that the aggressive optimization of in-domain capabilities does not lead to overfitting or "alignment tax." Instead, the multi-stage adaptation fosters a robust reasoning policy that generalizes effectively, balancing specialized optimization skills with broad reasoning transfer.

\subsection{NP-hard Tasks}
\subsubsection{Set Cover}
The task is to solve the classical Set Cover Problem. Given a universal set \( U \) and a collection of subsets \( S \subseteq 2^U \), the goal is to find the smallest possible sub-collection of \( S \) whose union equals \( U \). In other words, we aim to select the minimum number of subsets such that every element in \( U \) is contained in at least one of the selected subsets. If no such selection exists, the answer should be "Impossible". The solution is represented as a list of subset indices corresponding to the chosen sub-collection.

For example, given \( U = \{0,1,2,3,4,5\} \) and 
\begin{align*}
S = \{ & 0:\{0,1,2\},\; 1:\{2,3\},\; 2:\{0,4\}, \\
       & 3:\{3,4,5\},\; 4:\{1,2,5\} \},
\end{align*}
a valid minimum cover is \([0, 3, 4]\), since the union of these subsets is equal to \( U \).

The difficulty of the problem instances is categorized based on the size of the universe \( |U| \), the number of subsets \( |S| \), and the relative subset size (controlled by the parameter \(\text{subset\_size\_factor}\)):

\begin{itemize}
    \item \textbf{Easy}: 
    \begin{itemize}
        \item \( |U| \in [10, 20] \), \( |S| \in [5, 10] \), subset size factor \(= 0.4\)
        \item Small universe and relatively large subsets, making coverage straightforward.
    \end{itemize}
    
    \item \textbf{Medium}: 
    \begin{itemize}
        \item \( |U| \in [20, 25] \), \( |S| \in [10, 15] \), subset size factor \(= 0.4\)
        \item Moderate universe size and subset count, requiring careful selection.
    \end{itemize}
    
    \item \textbf{Hard}: 
    \begin{itemize}
        \item \( |U| \in [25, 30] \), \( |S| \in [15, 25] \), subset size factor \(= 0.4\)
        \item Larger universe with more subsets, increasing combinatorial complexity.
    \end{itemize}
    
    \item \textbf{Benchmark}: 
    \begin{itemize}
        \item \( |U| \in [30, 40] \), \( |S| \in [20, 30] \), subset size factor \(= 0.4\)
        \item The most challenging setting, with the largest universes and dense subset collections.
    \end{itemize}
\end{itemize}

\subsubsection{Subset Sum}
The Subset Sum Problem asks whether a subset of integers sums up to a given target value \( T \). In this variation, the objective is not only to reach the target sum, but also to maximize the number of elements used in the subset. 

Formally, given a set of integers \( \{a_0, a_1, \dots, a_{n-1}\} \) and a target \( T \), the task is to find an index set \( I \subseteq \{0,1,\dots,n-1\} \) such that 
\[
\sum_{i \in I} a_i = T,
\]
and among all valid solutions, the chosen \( I \) maximizes \(|I|\). If multiple such subsets exist, any of them is acceptable. The submission format requires returning the ordered list of indices, e.g., \([0,1,4]\).

For example, given \( T = 10 \) and 
\begin{align*}
\text{numbers} = \{ & 0:2,\; 1:3,\; 2:7, \\
                    & 3:8,\; 4:5 \},
\end{align*}
a valid solution is \([0,1,4]\), since \(2+3+5=10\), and the subset uses three elements, which is maximal.

The difficulty of generated problem instances is categorized according to the number of integers available (\(|\text{numbers}|\)), the typical size of the optimal solution (\(|I|\)), and the range of integer values:

\begin{itemize}
    \item \textbf{Easy}: 
    \begin{itemize}
        \item Total numbers \( \in [5, 10] \), solution size \( \in [4, 8] \), values in \([1, 5]\).
        \item Small input with low values, ensuring frequent feasible solutions.
    \end{itemize}
    
    \item \textbf{Medium}: 
    \begin{itemize}
        \item Total numbers \( \in [8, 12] \), solution size \( \in [4, 8] \), values in \([1, 10]\).
        \item Moderate instance size and range, requiring more careful subset selection.
    \end{itemize}
    
    \item \textbf{Hard}: 
    \begin{itemize}
        \item Total numbers \( \in [12, 15] \), solution size \( \in [8, 12] \), values in \([1, 15]\).
        \item Larger solution sizes and wider value ranges increase combinatorial difficulty.
    \end{itemize}
    
    \item \textbf{Benchmark}: 
    \begin{itemize}
        \item Total numbers \( \in [15, 20] \), solution size \( \in [10, 15] \), values in \([1, 15]\).
        \item The most challenging setting, with large search space and dense feasible solutions.
    \end{itemize}
\end{itemize}

\subsubsection{Knapsack}
The Knapsack Problem requires selecting a subset of items to maximize the total value without exceeding a weight capacity. Formally, given a set of items \( \{(w_i, v_i)\}_{i=0}^{n-1} \), each with weight \( w_i \) and value \( v_i \), and a knapsack capacity \( W \), the goal is to find an index set \( I \subseteq \{0,1,\dots,n-1\} \) such that
\[
\sum_{i \in I} w_i \leq W, \quad \text{and} \quad \sum_{i \in I} v_i
\]
is maximized. The submission format requires returning the ordered list of chosen item IDs in square brackets, e.g., \([0,2,3]\).

For example, with \( W = 20 \) and
\begin{align*}
\text{items} = \{ & 0:(3,4),\; 1:(4,5), \\
                  & 2:(7,10),\; 3:(8,11) \},
\end{align*}
a valid optimal solution is \([0,2,3]\), achieving total weight \(18 \leq 20\) and total value \(25\).

The problem instances are categorized into four difficulty levels, determined by the number of items, their weight/value ranges, and the relative knapsack capacity:

\begin{itemize}
    \item \textbf{Easy}: 
    \begin{itemize}
        \item \(6 \leq |I^*| \leq 10\) (solution items), total items \(\approx 15\text{--}25\).
        \item Weights in \([5,25]\), value-to-weight ratio in \([1.8,2.5]\).
        \item Capacity is \(1.1\text{--}1.4\) times the total weight of the solution items.
    \end{itemize}
    
    \item \textbf{Medium}: 
    \begin{itemize}
        \item \(8 \leq |I^*| \leq 12\), total items \(\approx 25\text{--}35\).
        \item Weights in \([20,80]\), value-to-weight ratio in \([1.5,2.0]\).
        \item Capacity is \(1.05\text{--}1.25\) times the total solution weight.
    \end{itemize}
    
    \item \textbf{Hard}: 
    \begin{itemize}
        \item \(15 \leq |I^*| \leq 25\), total items \(\approx 35\text{--}60\).
        \item Weights in \([50,200]\), value-to-weight ratio in \([1.2,1.6]\).
        \item Capacity is \(1.02\text{--}1.15\) times the total solution weight.
    \end{itemize}
    
    \item \textbf{Benchmark}: 
    \begin{itemize}
        \item \(25 \leq |I^*| \leq 35\), total items \(\approx 55\text{--}80\).
        \item Weights in \([50,200]\), value-to-weight ratio in \([1.2,1.6]\).
        \item Capacity is \(1.02\text{--}1.15\) times the total solution weight.
        \item The most challenging setting, with many items and tight capacity.
    \end{itemize}
\end{itemize}

\subsubsection{Balanced Minimum Bisection}
The Balanced Minimum Bisection Problem requires partitioning a weighted undirected graph \( G = (V, E) \) into two disjoint subsets of nearly equal size (differing by at most one vertex) such that the sum of the weights of edges crossing the cut is minimized. Unlike the classic Minimum Cut Problem, this task includes a balance constraint: both partitions must contain approximately the same number of vertices. 

Formally, let \( V \) be divided into \( V_1 \) and \( V_2 \) such that \( V_1 \cap V_2 = \emptyset \), \( V_1 \cup V_2 = V \), and \( \bigl||V_1| - |V_2|\bigr| \leq 1 \). The objective is to minimize
\[
\sum_{\substack{u \in V_1, v \in V_2 \\ (u,v) \in E}} w(u,v),
\]
where \( w(u,v) \) is the edge weight. The solution format specifies the two subsets explicitly, e.g., \([[0,1,2],[3,4,5]]\).

For example, consider the input graph:
\begin{align*}
0&: \{1:3,\, 2:1\}, \quad 1: \{0:3,\, 2:2,\, 3:2\}, \\
2&: \{0:1,\, 1:2,\, 3:3\}, \quad 3: \{1:2,\, 2:3\}.
\end{align*}
A valid optimal balanced bisection is \([[0,1],[2,3]]\).

The difficulty of generated instances is determined by the number of nodes, the structural complexity of the graph, and the noise level:

\begin{itemize}
    \item \textbf{Easy}:
    \begin{itemize}
        \item \(|V| \approx 30\).
        \item Graphs have clear community structures with dense intra-community edges and sparse inter-community connections, with small noise (\(\approx 0.1\)).
        \item Balanced cuts are relatively easy to identify.
    \end{itemize}
    
    \item \textbf{Medium}:
    \begin{itemize}
        \item \(|V| \approx 42\).
        \item Graphs exhibit fuzzier community boundaries and more inter-community edges, with moderate noise (\(\approx 0.15\)).
        \item Increases difficulty by reducing the clarity of the optimal partition.
    \end{itemize}
    
    \item \textbf{Hard}:
    \begin{itemize}
        \item \(|V| \approx 45\).
        \item Graphs are generated with deceptive structures, including “traitor” nodes and reinforced communities.
        \item Noise level around \(0.1\), making near-optimal but incorrect cuts more likely.
    \end{itemize}
    
    \item \textbf{Benchmark}:
    \begin{itemize}
        \item \(|V| \approx 50\).
        \item Graphs include reinforced “hell mode” structures, traitor nodes, and low noise (\(\approx 0.02\)).
        \item The most challenging setting, with multiple plausible partitions and high combinatorial complexity.
    \end{itemize}
\end{itemize}

\subsubsection{Meeting Scheduling Problem}
The Meeting Scheduling Problem (MSP) aims to assign meetings to rooms and times in order to maximize total attendee participation, subject to availability and capacity constraints. Each meeting requires a set of attendees and a duration, each attendee has availability intervals, and each room has a capacity. A feasible solution must assign to each scheduled meeting a start time and a room such that:
\begin{itemize}
    \item All required attendees are available for the entire duration.
    \item The room has sufficient capacity for all attendees.
    \item No attendee or room is scheduled for overlapping meetings.
\end{itemize}
If a meeting cannot be scheduled under these constraints, it is omitted. The solution is expressed as an ordered list of tuples \((\text{meeting\_id}, \text{room\_id}, \text{start\_time})\), sorted by start time.

For example, given the input:
\begin{align*}
\text{meetings} = \{ & 0:([0,1,2],60),\; 1:([1,3],30), \\
                     & 2:([0,2,3],90) \}, \\
\text{availability} = \{ & 0:[(900,1700)], \\
                         & 1:[(900,1200),(1300,1700)], \\
                         & 2:[(900,1700)],\; 3:[(1000,1400)] \}, \\
\text{rooms} = \{ & 0:5,\; 1:3 \},
\end{align*}
a valid schedule is
\[
[(0,0,900),\; (1,1,1000),\; (2,0,1020)],
\]
which yields a total of 8 attendee participations.

The difficulty of generated MSP instances depends on the number of meetings, attendees, rooms, and fragmentation of availability:

\begin{itemize}
    \item \textbf{Easy}:
    \begin{itemize}
        \item 4--5 meetings, 3--5 attendees, 3--4 rooms.
        \item At most 3 attendees per meeting.
        \item Availability mostly continuous within the working day.
    \end{itemize}

    \item \textbf{Medium}:
    \begin{itemize}
        \item 5--6 meetings, 4--6 attendees, 4--5 rooms.
        \item At most 4 attendees per meeting.
        \item Some attendees have fragmented availability (e.g., lunch breaks).
    \end{itemize}

    \item \textbf{Hard}:
    \begin{itemize}
        \item 6--7 meetings, 5--7 attendees, 5--6 rooms.
        \item At most 4 attendees per meeting.
        \item Heavier overlap among meetings and tighter room capacities.
    \end{itemize}

    \item \textbf{Benchmark}:
    \begin{itemize}
        \item 8--10 meetings, 7--9 attendees, 6--7 rooms.
        \item At most 5 attendees per meeting.
        \item The most challenging setting, with dense scheduling conflicts and fragmented availability.
    \end{itemize}
\end{itemize}

\subsubsection{Hamiltonian cycle}
The task is to find a Hamiltonian circuit in a given graph \( G \), which is a path that visits every vertex exactly once and returns to the starting point. The goal is to maximize the number of vertices included in the Hamiltonian circuit. The process starts with a random vertex and finds a small valid subgraph, then iteratively expands the subgraph while ensuring it remains valid, continuing until the largest possible Hamiltonian circuit is found.

The problem is categorized into four difficulty levels based on the number of vertices and edge density:

\begin{itemize}
    \item \textbf{Easy}: 
    \begin{itemize}
        \item \( |V| \in [15, 20] \), \( \rho = 0.2 \)
        \item Small graph with sparse edges.
    \end{itemize}
    
    \item \textbf{Medium}: 
    \begin{itemize}
        \item \( |V| \in [20, 30] \), \( \rho = 0.3 \)
        \item Moderate graph size with moderate connectivity.
    \end{itemize}
    
    \item \textbf{Hard}: 
    \begin{itemize}
        \item \( |V| \in [30, 40] \), \( \rho = 0.4 \)
        \item Larger graph with denser edges, increasing difficulty.
    \end{itemize}
    
    \item \textbf{Benchmark}: 
    \begin{itemize}
        \item \( |V| \in [40, 50] \), \( \rho = 0.5 \)
        \item The most challenging, with the largest and densest graph.
    \end{itemize}
\end{itemize}

\subsubsection{Traveling Salesman Problem}
The Traveling Salesman Problem (TSP) is a classical combinatorial optimization problem. Given a set of cities and pairwise distances, the objective is to find the shortest possible tour that:
\begin{itemize}
    \item Starts and ends at the same city.
    \item Visits each city exactly once in between.
\end{itemize}

The solution is expressed as a route \([c_0, c_1, \dots, c_{n-1}, c_0]\), where \(c_0\) is the starting city and each city appears exactly once except for the repetition of \(c_0\) at the end.

For example, given the distance dictionary:
\begin{align*}
0: \{1:10,\,2:15,\,3:20\}, \\ 1: \{0:10,\,2:35,\,3:25\}, \\
2: \{0:15,\,1:35,\,3:30\}, \\ 3: \{0:20,\,1:25,\,2:30\},
\end{align*}
a valid optimal solution is
\[
[0,1,3,2,0].
\]
The difficulty of generated TSP instances is determined primarily by the number of cities:
\begin{itemize}
    \item \textbf{Easy}: 10--20 cities.
    \item \textbf{Medium}: 20--30 cities.
    \item \textbf{Hard}: 35--45 cities.
    \item \textbf{Benchmark}: 45--55 cities.
\end{itemize}
All instances are generated with symmetric distance matrices, with distances sampled uniformly within a predefined range.

\subsubsection{Maximum Clique Problem}
The Maximum Clique Problem (MCP) is defined on an undirected graph \(G = (V,E)\).  
A clique is a subset of vertices \(C \subseteq V\) such that every pair of distinct vertices in \(C\) is connected by an edge in \(E\).  
The problem asks for the largest such subset, i.e., a clique of maximum cardinality.

The solution is expressed as a list of vertex IDs forming the clique.  
For example, given the adjacency lists:
\begin{align*}
0&: [1,2,3,4], \quad 1: [0,3,4], \quad 2: [0,3], \\
3&: [0,1,2,4], \quad 4: [0,1,3],
\end{align*}
a valid maximum clique is
\[
[0,1,3,4],
\]
which has size 4.

The difficulty of generated MCP instances depends on the graph size and density:
\begin{itemize}
    \item \textbf{Easy}: 4--8 vertices, cliques of size 2--4.  
    \item \textbf{Medium}: 8--12 vertices, cliques of size 2--4.  
    \item \textbf{Hard}: 12--16 vertices, cliques of size 2--6.  
    \item \textbf{Benchmark}: 16--20 vertices, cliques of size 4--8.  
\end{itemize}
Graphs are generated by first constructing a guaranteed clique and embedding it into a larger graph with random edges, ensuring the clique exists as the maximum solution.

\subsubsection{Maximum Independent Set}
The Maximum Independent Set (MIS) problem is defined on an undirected graph \(G = (V,E)\).  
An independent set is a subset of vertices \(I \subseteq V\) such that no two vertices in \(I\) are adjacent in \(G\).  
The problem asks for the independent set of maximum cardinality.

The solution is expressed as a list of vertex IDs forming the set.  
For example, given the adjacency lists
\[
0:\{1,2\},\quad
1:\{0,2,3\},\quad
2:\{0,1,3\},\quad
3:\{1,2\},
\]
a maximum independent set is
\[
[0,3],
\]
which has size 2.

The difficulty of generated MIS instances depends mainly on the graph size and the planted independent set:
\begin{itemize}
    \item \textbf{Easy}: 12--20 vertices, independent set size 4--8.
    \item \textbf{Medium}: 20--30 vertices, independent set size 8--12.
    \item \textbf{Hard}: 30--40 vertices, independent set size 12--16.
    \item \textbf{Benchmark}: 40--50 vertices, independent set size 16--20.
\end{itemize}
Graphs are generated by first selecting a guaranteed independent set and embedding it into a larger graph with randomly added edges, ensuring the independent set exists as the maximum solution.

\subsubsection{Graph Coloring Problem}
The Graph Coloring Problem (GCP) is defined on an undirected graph \(G=(V,E)\).  
The task is to assign a color to each vertex such that no two adjacent vertices share the same color, while minimizing the total number of colors used.

The solution is expressed as a list of integers, where the \(i\)-th entry denotes the color assigned to vertex \(i\).  
For example, given the adjacency lists
\[
0:[1,2],\quad
1:[0,3],\quad
2:[0,3],\quad
3:[1,2],
\]
a valid optimal coloring is
\[
[1,2,1,2],
\]
which uses 2 colors.

The difficulty of generated GCP instances depends mainly on the number of vertices, the number of colors required, and the edge density:
\begin{itemize}
    \item \textbf{Easy}: 8--12 vertices, 3--4 colors, edge density \(\approx 0.2\).
    \item \textbf{Medium}: 15--22 vertices, 4--6 colors, edge density \(\approx 0.35\).
    \item \textbf{Hard}: 25--32 vertices, 6--8 colors, edge density \(\approx 0.5\).
    \item \textbf{Benchmark}: 32--40 vertices, 6--8 colors, edge density \(\approx 0.5\).
\end{itemize}
Graphs are generated by partitioning vertices into color classes and adding random edges between different partitions, ensuring that the planted coloring remains a valid optimal solution.


\end{document}